\documentclass[11pt]{article}
\usepackage{acl2016}
\usepackage{amsmath}
\usepackage{times}
\usepackage{url}
\usepackage{morefloats}
\usepackage{graphicx}
\usepackage{pgfplots}
\pgfplotsset{compat=1.3}
\usetikzlibrary{shapes,shadows,positioning,calc,patterns}
\usepackage{multirow}
\usepackage{picinpar}
\usepackage{changepage}
\usepackage{epstopdf}
\usepackage{breqn}

\usepackage[ruled,vlined]{algorithm2e}

\def\sm{\small}
\def\scrs{\scriptsize}

\DeclareMathOperator{\Ob}{\mathbf{O}}
\DeclareMathOperator{\ab}{\mathbf{a}}
\DeclareMathOperator{\tb}{\mathbf{t}}
\DeclareMathOperator{\tpb}{\mathbf{t}^\prime}
\DeclareMathOperator{\ub}{\mathbf{u}}
\DeclareMathOperator{\wb}{\mathbf{w}}
\DeclareMathOperator{\thetab}{\mathbf{\Theta}}
\newcommand{\cbr}[1]{\left\{#1\right\}}

\def\spairta{\langle \tb, \ab \rangle}
\def\spairtpa{\langle \tpb, \ab \rangle}
\def\spairtpt{\langle \tpb, \tb \rangle}
\def\pv{\textsc{PV}}
\def\mcc{\textsc{MC$_C$}}
\def\mcafp{\textsc{MC$_{\textsc{\scrs AFP}}\;$}}
\def\mcafpr{\textsc{MC$_{\textsc{\scrs AFP}}^{\mbox{\scrs Rnd}}\;$}}
\def\simsurf{\textsf{sim}_{\small \textsc{surf}}}
\def\simpv{\textsf{sim}_{\small \pv}}
\def\ffnnf{FFNN$_{\mbox{\scrs 5-class}}\;$}
\def\ffnnb{FFNN$_{\mbox{\scrs 2-class}}^{\mbox{\scrs argmax 1..5}}\;$}
\def\seqff{Seq2seq+FFNN$\;$}

\aclfinalcopy

\title{Building Large Machine Reading-Comprehension Datasets using Paragraph Vectors}

\author{
  Radu Soricut \\
  Google Inc. \\
  {\tt rsoricut@google.com} \\\And
  Nan Ding \\
  Google Inc. \\
  {\tt dingnan@google.com} \\}

\date{}

\begin{document}
\maketitle
\begin{abstract}
We present a dual contribution to the task of machine reading-comprehension:
a technique for creating large-sized machine-comprehension (MC) datasets using paragraph-vector models;
and a novel, hybrid neural-network architecture that combines the representation power
of recurrent neural networks with the discriminative power of fully-connected
multi-layered networks.
We use the MC-dataset generation technique to build a dataset of around 2 million
examples, for which we empirically determine the high-ceiling of human performance (around 91\% accuracy),
as well as the performance of a variety of computer models.
Among all the models we have experimented with, our hybrid neural-network architecture achieves
the highest performance (83.2\% accuracy).
The remaining gap to the human-performance ceiling provides enough room for future model improvements.
\end{abstract}

\section{Introduction}
The ability to have computer models that can achieve genuine understanding of natural language text
is a long-standing goal of Artificial Intelligence~\cite{norvig:1978}.
The problem can be formulated in a relatively straight-forward manner, by borrowing the same techniques
that are used for measuring reading comprehension in human students:
given a text passage, measure how well they can answer questions related to the information being presented
in that passage.

Previous attempts towards achieving this goal have ranged from hand-engineered grammars~\cite{riloff-thelen:2000},
to predicate-argument extraction and querying~\cite{poon-etal:2010}, to exploiting
recently-proposed reading-comprehension datasets~\cite{richardson-etal:2013,berant-etal:2014,hermann-etal:2015,rajpurkar-etal:2016}
using various machine learning approaches~\cite{wang-etal:2015,yin-etal:2016,chen-etal:2016,sordoni-etal:2016}.
To better understand the current state of affairs in this field, let us take a look at two
publicly-available reading-comprehension datasets and recap the current state-of-the-art (SOTA) results.

The MCTest~\cite{richardson-etal:2013} contains about 500 stories associated with 2000 questions
(4 questions per story), and 4 choices per question (only one correct);
the stories and questions are written at the reading-level of elementary-school students
(e.g., ``James the Turtle was always getting in trouble.'').
The human-level performance on this dataset is estimated to be in the high 90\% range.
The SOTA on this dataset is about 70\% accuracy~\cite{wang-etal:2015}, obtained using a model
that combines both traditional and more recent language-understanding tools,
such as syntax, frame semantics, co-reference, and word embeddings.
It's worth noting here that models using various neural-network architectures achieve
only about 53\% accuracy~\cite{yin-etal:2016},
presumably because the small data size does not allow for optimal parameter estimation.

At the other end of the spectrum is the CNN/DailyMail dataset~\cite{hermann-etal:2015}, which
is based on news articles from CNN and DailyMail; the number of examples are in the hundreds of thousands,
and the reading-level of these text is high-school/college level;
the 'questions' are created from the bullet-point summaries that accompany these articles, and
a clever entity-anonimization technique is used to ensure that only the information from the
current passage is used to provide the answer
(e.g., ``According to the sci-fi website @entity9 , the upcoming novel @entity11 will feature a capable but flawed @entity13 official ...'').
On these datasets, the human-level performance is estimated to be around 75\% accuracy~\cite{chen-etal:2016},
while the SOTA is currently at 76.1 on CNN~\cite{sordoni-etal:2016}, and 75.8 on DailyMail~\cite{chen-etal:2016}.

Consistent with the last research findings in a variety of natural-language tasks for which enough
supervised training data is available, the SOTA results are achieved using
neural-network models.
For instance, the results in~\cite{chen-etal:2016} are achieved using a neural-network architecture
based on the {\it AttentiveReader} model originally proposed by~\cite{hermann-etal:2015},
together with some modifications that prove effective for the task.
But since SOTA and human accuracy are so close for the CNN/DailyMail dataset, we conclude that it has fulfilled its mission
and is now headed towards retirement (at least in the current form as a machine-comprehension dataset;
see~\cite{nallapati-etal:2016} for a useful reincarnation in the context of abstractive summarization).

In this context, we present a dual contribution to the field of Machine Comprehension of text.
First, we present a technique via which one can obtain machine-comprehension datasets based on
widely-available news data (e.g., the English Gigaword~\cite{ldc-english-gigaword});
we use this technique to generate a datasets of around 2 million examples,
on which we estimate that the human-level accuracy is in the 90\% range
(in a 5-way multi-choice setup; for comparison, a random-guess approach has 20\% accuracy).
Second, we present a novel neural-network architecture that combines the representation power
of recurrent neural networks with the discriminative power of fully-connected multi-layered
networks, and achieves the best results we could obtain on our dataset: 83.2\% accuracy.

These contributions open the doors for building interesting challenges for the
Machine Comprehension task, along the following dimensions:
we build such datasets over text that uses rich language constructs
on both the input side (hence a requirement for understanding the context)
and the output side (hence a need for understanding the answers),
thus capturing a large spectrum of natural language understanding challenges;
the large size of such datasets allows for exploring the performance of machine-learning models
in a supervised-learning setting, including data-hungry models such as deep neural-network architectures;
last but not least, we can use the difference between the human-performance ceiling on these datasets and
the current SOTA to assess the progress made in the field of Machine Comprehension of text.
We release the machine-comprehension dataset used in this paper to the research community, in order to facilitate
direct comparisons against the results reported in this paper, as well as boost the influx of
new contributions to the field.

\section{Building a Paragraph-Vector--based MC Dataset}
In this section, we describe an algorithm for the creation of Machine Comprehension datasets,
based on exploiting Paragraph Vector~\cite{le-mikolov:2014} models.
The data requirement is that of a large set $C$ of
$\langle \mbox{\it title}, \mbox{\it article}\rangle$ pairs,
of the type that can be extracted from a variety of publicly-available corpora,
for instance the English Gigaword~\cite{ldc-english-gigaword}.

\subsection{The \textsc{MC-Create} Algorithm}
\label{sec:alg}
The pseudo-code for the creation of the dataset is presented below:

\begin{algorithm}
\caption{\textsc{MC-Create}($C$, $N$, $\mbox{\it Score}$)}
\SetAlgoLined
\KwResult{Dataset \mcc}
Train paragraph-vector model $\textsf{PV}_C$ \\
\mcc$\gets \emptyset$ \\
$nr\_decoys = 4$ \\
\For{$\spairta \in C$}{
  $A\gets []$ \\
  $T_t \gets \mbox{\it PV}_C(\tb)[1..N]$ \\
  \For{$\tpb\in T_t$}{
    $score \gets Score(\tpb, \spairta)$ \\
    \If{$score > 0$}{
        $A.\mbox{\bf append}(\langle score, \tpb\rangle)$
      }
    }
  \If{$|A| \geq nr\_decoys$}{
      \mcc$\gets$ \mcc$\cup\{(\spairta, \mbox{\bf true})\}$ \\
      $R\gets \mbox{\bf descending-sort}(A)$ \\
      \For{$i \in [1..nr\_decoys]$}{
        $\langle score, \tpb\rangle\gets R[i]$ \\
        \mcc$\gets$ \mcc$\cup\{(\spairtpa, \mbox{\bf false})\}$ \\
      }
    }
}
\end{algorithm}

Algorithm \textsc{MC-Create} takes three main parameters:
a dataset $C = \{ \langle \tb_i, \ab_i \rangle | 1 \leq i \leq m \}$;
an integer $N$ that controls the paragraph-vector neighborhood size;
and a function $\mbox{\it Score}$, used to score the $N$ entries for each such neighborhood.

The first step of the algorithm involves training a paragraph-vector model using the title
entries in dataset $C$.
For a given title $\tb$ in $C$, its $N$ closest neighbors in $PV_C$ are denoted $T_t$.
Each title entry $\tpb$ in $T_t$ is scored against the $\spairta$ pair according to the
$\mbox{\it Score}$ function.
In what follows, we are using the following $\textsf{Score}$ function definition:

\begin{dmath}
\textsf{Score}(\tpb, \spairta) =
\left\{
\begin{array}{rl}
0, & {\scriptstyle \mbox{if}\, \simsurf (\tpb, \tb) \geq L} \\
\lambda_e \simpv(\tpb, \tb)   & \\
+ \lambda_s \simsurf(\tpb, \tb)    & {\scriptstyle\mbox{otherwise}}\\
+ (1-\lambda_s) \simsurf(\tpb, \ab))& \\
\end{array}
\right.
\label{eq:score}
\end{dmath}
\noindent
This function uses the notation $\simsurf(\tpb, \tb)$ to indicate a similarity
score based on the surface of the strings $\tpb$ and $\tb$,
and $\simpv(\tpb, \tb)$ to indicate a similarity
score based on the paragraph-vector representations of $\tpb$ and $\tb$.

The $\mbox{\it Score}$ function first verifies if the surface-based similarity for pair
$\spairtpt$ exceeds some threshold $L$.
In that case, the function returns 0, as a guard against considering pairs that are too surface-similar;
otherwise, it computes a weighted linear combination (using weights $\lambda_e$ and $\lambda_s$)
between the embedding-based similarity $\simpv(\tpb, \tb)$ and the
surface-based similarities $\simsurf$ for both $\spairtpt$ and $\spairtpa$.
Intuitively, the higher this value is, the more likely it is that $\tpb$ is a good
decoy for $\tb$ with respect to article $\ab$.

For the dataset released with this paper and the experiments contained here,
we used the following hyper-parameters:
$\simsurf(\tpb, \tb)$ is computed using BLEU~\cite{papineni-etal:2002} (with brevity-penalty=1);
$\simpv(\tpb, \tb)$ is computed as the cosine between the embedding representations
\pv$(\tpb)$ and \pv$(\tb)$;
$\lambda_e$ = 1; $\lambda_s$ = 0.5; $L=0.5$; the paragraph-vector model \pv$_C$ is trained
using the PV-DBOW model~\cite{le-mikolov:2014} using softmax sampling,
for 5 epochs with a token minimum count of 5, and an embedding size of 256.

\begin{table}[th]
\begin{center}
\begin{tabular}{lrrr}
 Stats                     & Train     &  Dev  & Test \\ \hline
\#Instances   & 1,727,423 & 7,602 & 7,593  \\
Avg. tokens/article        &    93.7   & 96.4   & 96.8 \\
Avg. tokens/answer         &     9.6   &  9.9   &  9.9 \\
\end{tabular}
\caption{Statistics over the dataset created by \textsc{MC-Create} over the AFP portion of the Gigaword.}
\label{tab:mcafp-stats}
\end{center}
\end{table}

\begin{table*}[th]
\begin{center}
\begin{tabular}{lll}
Article & Options \\ \hline
\sm Two elite Australian air-force pilots were suspended for & \sm 1. Australian pilots suspended over landing gear miss: officials \\
\sm going out drinking the night before they were were due to & \sm 2. Cambodian PM's nephew cleared over fatal shootings, released \\
\sm fly Prime Minister Julia Gillard, the defence department & \sm 3. Britain to hand over Basra in two weeks: PM's office \\
\sm said on Friday. The two members of the Royal Australian  & \sm 4. Aussie PM's pilots suspended over drinking spree \\
\sm Air Force's 34 Squadron were temporarily suspended but   & \sm 5. Foreign media in a tangle over Spanish PM's name \\
\sm not disciplined further after the incident, the department & \\
\sm said.                                                      & \\ \hline
\sm Israeli Foreign Minister Avigdor Lieberman's UN General & \sm 1. Climate czar's departure will not impact talks: UN official\\
\sm Assembly speech, which outlined controversial proposals & \sm 2. Iraq "doesn't understand" UN demands: Powell \\
\sm for an Israeli-Palestinian agreement, did not reflect the & \sm 3. Israeli FM to hold talks with Rice, UN chief Annan on trip to US \\
\sm official Israeli position, the premier's office said Tuesday. & \sm 4. Lieberman UN speech doesn't reflect Israel view on talks: PM \\
\sm Prime Minister Benjamin Netanyahu appeared to be distan- & \sm 5. Syria "not pessimistic" on peace with Israel: Assad \\
\sm cing himself from Lieberman's more controversial proposals. & \\ \hline
\sm The United States is pushing for peace talks between & \sm 1. Israel, Syria not ready for peace talks: US envoy \\
\sm Israel, Syria and Lebanon, US envoy George Mitchell & \sm 2. Israel, Syria to resume stalled peace talks: US \\
\sm said Tuesday, as the Israelis prepared to resume     & \sm 3. Israel, Syria want to resume peace talks: US \\
\sm direct negotiations with the Palestinians. & \sm 4. Lebanon, Syria discuss peace talks with Israel \\
                                               & \sm 5. US seeks Israeli peace talks with Syria, Lebanon: envoy \\ \hline
\end{tabular}
\caption{Examples of MC instances created by \textsc{MC-Create} over the AFP portion of the Gigaword.}
\label{tab:mcafp-examples}
\end{center}
\end{table*}

\begin{table*}[ht]
  \begin{center}
  \begin{tabular}{ c ccc cccc}
             & \multicolumn{7}{c}{Dev dataset (sample)} \\ \cline{2-7}
             & \multicolumn{3}{c}{Difficulty} & \multicolumn{4}{c}{Accuracy} \\
      Method & Easy & Medium & Hard & Easy & Medium & Hard & Overall  \\ \hline
      Human  & 81\% &  14\%  & 5\%  & 93.6 &  86.4  & 57.1 & 90.9 \\ \cline{1-1}\cline{5-8}
      BLEU (brev-penalty=1)   & & & & 55.9 &  38.2  & 38.9 & 51.0 \\
      Paragraph-Vectors       & & & & 16.8 &  14.7  &  5.6 & 15.0 \\
      Random                  & & & & 19.6 &  20.6  & 16.7 & 19.5  \\
  \end{tabular}
  \end{center}
  \caption{Performance range for human and baseline methods on the AFP machine comprehension task.}
  \label{tab:range}
\end{table*}

\subsection{Examples and Performance Range}
We use the procedure described in Section~\ref{sec:alg} to generate a machine-comprehension dataset,
using as $C$ the portion of the Gigaword corpus~\cite{ldc-english-gigaword} restricted to
articles published by AFP (Agence France-Presse).
The reason we restrict to one publisher has to do with information overlap: for a given news event,
multiple publishers write headline \& article pairs on that event, often to the extent that a headline
from publisher A can fit perfectly an article from publisher B.
Since this effect would muddle the one-correct-answer assumption for our dataset, we therefore recommend
restricting the news corpus used to one publisher.
The same effect can exist, to a smaller extent, even for a one-publisher corpus; we mitigate against it at a surface level,
using threshold $L$ in the $\mbox{\it Score}$ function (Equation~\ref{eq:score}) used by the \textsc{MC-Create} algorithm.
We also note here that the size of set $C$ plays an important role in the creation of a challenging dataset;
in our experience, it needs to be on the order of a few million examples, with smaller set sizes resulting in less challenging datasets.

The performance metric for such a dataset is straightforward:
we compute accuracy by comparing the extent to which,
when presented with an entry $E = \{ \langle \tb_i^j, \ab_i \rangle | 1 \leq j \leq 5 \}$,
the choice for index $j^*$, $1 \leq j^* \leq 5$,
coincides with having $(\langle \tb_i^{j^*}, \ab_i \rangle, \mbox{\bf true})$ in \textsc{MC$_C$}.

We denote the MC dataset resulted from restricting to the AFP portion of the Gigaword as the \mcafp set.
This dataset, including the train/dev/test splits used in the experiments reported in Section~\ref{sec:experiments},
is available online\footnote{{\tt http://www.github.com/google/mcafp}}.
(Note that you still need to have access to LDC's English Gigaword Fifth Edition corpus~\cite{ldc-english-gigaword},
in order to be able to generate the \mcafp dataset.)
In Table~\ref{tab:mcafp-stats} we present some statistics over the train/dev/test splits\footnote{
Average token statistics computed over {\it tokenized} text.} for the \mcafp set.
Table~\ref{tab:mcafp-examples} contains examples extracted from this dataset.

When considering the examples in Table~\ref{tab:mcafp-examples},
we note that the 'Article' and 'Options' columns illustrate a large spectrum
of natural language understanding challenges:
from simple lexical matching (e.g., 'PM' and 'Prime Minister'),
to complex lexicon usage (e.g., 'drinking spree'),
to {\it who-did-what-to-whom} type of relationship determination.
Our conjecture is that solving the challenge posed by such a dataset
requires sophisticated language-understanding mechanisms.

To that end, we evaluate the performance of human annotators, as well as baseline algorithms,
on a sample of size 200 extracted from this dataset, see Table~\ref{tab:range}.
The human evaluators were asked, in addition to providing their best guess for the correct choice,
to subjectively assess whether they perceived each example as 'Easy', 'Medium', or 'Hard'.
For instance, the first example in Table~\ref{tab:mcafp-examples} was annotated as 'Easy',
the second example 'Medium', and the last example 'Hard'.
It is interesting to note that the human accuracy scores accurately reflect this subjective difficulty assessment
(first row in Table~\ref{tab:range}),
with 'Easy' examples averaging 93.6\% accuracy, 'Medium' at 86.4\%, and 'Hard' at 57.1\%, for
an overall performance of the human evaluator at 90.9\% accuracy.

The surface-based baseline BLEU (brevity-penalty=1) chooses the maximum-scoring option using the
BLEU~\cite{papineni-etal:2002} metric.
Its overall accuracy on this sample is 51.0\%.
This metric's accuracy when considering the human-created labels of 'Easy', 'Medium', and 'Hard' also
reflects the fact that perceived difficulty correlates with surface string matching,
with 'Easy' examples averaging 55.9\% accuracy, while 'Medium' and 'Hard' ones averaging around 38\% accuracy.

The other non-trivial metric, Paragraph-Vectors, chooses the maximum-scoring option as the maximum
cosine score between the article and each option.
Because of its high correlation with how Algorithm \textsc{MC-Create} computes the decoy options,
its performance at 15\% accuracy is actually below that of random guessing (19.5\% on this sample,
expected to be 20\% due to the 5 choices involved).

\section{Neural Network--based MC Models}
In this section, we present several neural-network--based models that we use
to tackle the challenge presented by \mcafp.
We use the following notations to describe these models:
each training instance pair is a tuple $\langle \tb_i, \ab_i^j \rangle$,
where $\ab_i$ denotes the article, and $\tb_i^j$ denotes the title option;
in addition, we use a binary variable $y_{ijk} \in \{0,1\}$ to denote whether
the $j$-th title of instance $i$ is labeled as $k$, and $\sum_k y_{ijk} = 1$.

\subsection{Feedforward Neural Network Models}
\label{sec:ffnn}
We experiment with the ability of standard feedforward neural-network models
to solve the \mcafp task.

We consider two types of classifiers. A 2-class--classifier (1 for 'yes', this is a correct answer; 0 for 'no', this is an incorrect answer) applied independently on all $\langle \tb_i, \ab_i^j \rangle$ pairs; the final prediction is the title with the highest 'yes' probability among all instance pairs belonging to instance $i$.

For each instance pair $\langle \tb_i, \ab_i^j \rangle$, the input to the neural network is an embedding
tuple $\langle \text{Emb}(\ab_i; \Omega), \text{Emb}(\tb_i^j; \Omega) \rangle$,
where $\Omega$ is the embedding matrix, and
$\text{Emb}(.)$ denotes the mapping from a list of word IDs to a list of
embedding vectors using $\Omega$.

Using these notations, the loss function for an FFNN can be written as:
\begin{dmath}
L(\Omega,\ub) = \sum_{i,j,k} y_{ijk} \log \; \text{FN}_k(\text{Emb}(\ab_i; \Omega), \text{Emb}(\tb_i^j; \Omega); \ub)
\label{eq:closs}
\end{dmath}
\noindent
where $\text{FN}_k$ denotes the $k$-th output of a feedforward neural network,
and $\sum_k \text{FN}_k(.) = 1$.
Our architecture uses a two hidden-layer fully connected network
with Rectified Linear hidden units, and a softmax layer on top.

In addition, we also consider a multi-class classifier, where each class $k$ corresponds to the index of the answer choice. In this case, we slightly modify Eq. \ref{eq:closs} by
\begin{dmath*}
L(\Omega,\ub) = \sum_{i,k} y_{ik} \log \; \text{FN}_k(\text{Emb}(\ab_i; \Omega), \text{Emb}(\tb_i^:; \Omega); \ub)
\end{dmath*}
\noindent
where $\tb_i^: = \cbr{\tb_i^1, \ldots, \tb_i^5}$ is a concatenation of all 5 candidate titles of instance $i$.

\subsection{Seq2seq + Feedforward Models}
\label{sec:seq+ffnn}
\begin{figure}[th]
\centering
\includegraphics[width=3in]{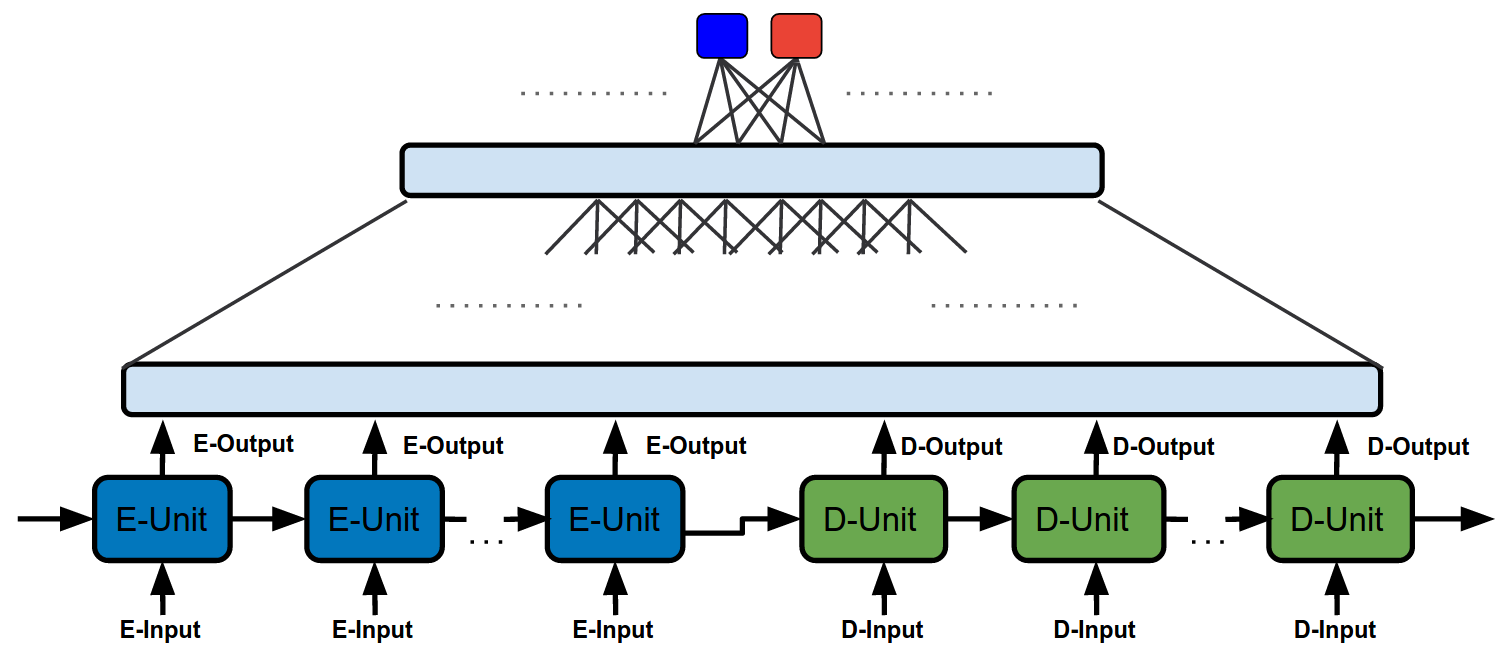}
\caption{Seq2seq + FFNN model architecture.}
\label{fig:agmc_diagram}
\end{figure}

We describe here a novel, hybrid neural-network model that combines a recurrent neural-network
with a feedforward one.
We note that both strings in a tuple $\langle \tb_i, \ab_i^j \rangle$ can be modeled as the encoding and decoding sequences
of a sequence-to-sequence (Seq2seq) model~\cite{sutskever-etal:2014,bahdanau-etal:2015}.
The output of each unit cell of a Seq2seq model (both on the encoding side and the decoding side)
can be fed into an FFNN architecture for binary classification.
See Figure~\ref{fig:agmc_diagram} for an illustration of the Seq2seq + FFNN model architecture.

In addition to the classification loss (Equation~\ref{eq:closs}), we also include a loss for generating an output sequence
$\tb_i^j$ based on an input $\ab_i$ sequence.
We define a binary variable $z_{ijlv} \in \{0,1\}$ to indicate whether
the $l$th word of $\tb_i^j$ is equal to word $v$.
$\Ob^d_{ijl}$ denotes the $l$-th output of the decoder of $\langle \tb_i, \ab_i^j \rangle$,
$\Ob^e_{ij:}$ denotes the concatenation of all the output of the encoder
of $\langle \tb_i, \ab_i^j \rangle$, and $\Ob^d_{ij:}$ denotes the concatenation of decoder outputs of $\langle \tb_i, \ab_i^j \rangle$.

With these definitions, the loss function for the Seq2seq + FFNN model is:
\begin{dmath}
L(\thetab, \wb, \ub) =
\sum_{i,j,k} y_{ijk} \log \; \text{FN}_k(\Ob^e_{ij:}(\ab_i, \tb_i^j; \thetab), \Ob^d_{ij:}(\ab_i, \tb_i^j; \thetab); \ub)
+ \lambda_{gen} \sum_{i,j,l,v} y_{ij1} z_{ijlv} \log \;\text{softmax}_v(\Ob^d_{ijl}(\ab_i, \tb_i^j; \thetab); \wb)
\label{eq:mloss}
\end{dmath}
\noindent
where $\sum_v \text{softmax}_v(.) = 1$;
$\thetab$ are the parameters of the Seq2seq model,
which include the parameters within each unit cell,
as well as the elements in the embedding matrices for source and target sequences;
$\wb$ are the output projection parameters that transform the output space of the decoder
to the vocabulary space for abstract generation;
$\ub$ are the parameters of the FFNN model (Equation~\ref{eq:closs});
$\lambda_{gen}$ is the weight assigned to the sequence-to-sequence generation loss.
Only the true target candidates (the ones with $y_{ij1} = 1$) are included in this loss,
as we do not want the decoy target options to affect this computation.

The Seq2seq model we use here is the attention-enhanced models
proposed in~\cite{bahdanau-etal:2015,chen-etal:2016}.
We apply Gated Recurrent Unit (GRU) as the unit cell~\cite{cho-etal:2014}.
For comparison purposes, the attention probabilistic mask $\alpha$ is instantiated using either
the tanh layer as in~\cite{bahdanau-etal:2015}, or the bilinear term as in~\cite{chen-etal:2016}.
We also perform ablation tests to measure the impact on performance of the $\lambda_{gen}$
parameter in Section~\ref{sec:lambda_gen}, and ablation tests to compare the performance of tied/untied
embedding matrix for encoding and decoding sequences in Section~\ref{sec:ablation-tied}.

\section{Experiments}
\label{sec:experiments}
\subsection{Experimental Setup}
The experiments with neural network models are done using the Tensorflow package~\cite{tensorflow2015-whitepaper}.
The hyper-parameter choices are decided using the hold-out development portion of the \mcafp set.
For modeling the input tokens, we use a vocabulary size of 100,000 types
(selected as the most frequent tokens over the titles of the AFP portion of the English Gigaword corpus~\cite{ldc-english-gigaword}),
with an embedding size of 512.
The models are optimized using ADAGRAD with a learning rate of 0.01, and clipped gradients (maximum norm 4).
We run the training procedure for 1M steps, with a mini-batch size of 200.
We use 40 workers for computing the updates, and 10 parameter servers for model storing and (asynchronous and distributed) updating.

We use the following notations to refer to the neural network models:
\ffnnf refers to a feedforward neural network architecture with a 5-class--classifier, each class corresponding to an answer choice;
we also refer to this model as FFNN5.
\ffnnb refers to a feedforward neural network architecture with a 2-class--classifier ('yes' or 'no' for answer correctness),
over which an argmax function computes a 5-way decision (i.e., the choice with the highest 'yes' probability);
we henceforth refer to this model simply as FFNN.
Both models use a two-hidden--layer architecture with sizes 1024 and 256.

The \seqff refers to the hybrid model described in Section~\ref{sec:seq+ffnn}, in which the Seq2seq model is combined with the \ffnnb model.
The RNN part of the model uses a two-hidden--layer GRU unit-cell configuration, while the FFNN part uses a two-hidden--layer architecture with sizes  64 and 16.
(We note here that for the hybrid model we also tried a larger FFNN setup with hidden layers of sizes 1024 and 256, but did not see improvements.)

The $\lambda_{gen}$ hyper-parameter from the loss-function $L(\thetab, \wb, \ub)$ (Equation~\ref{eq:mloss}) is by default set to 0.01
(except for Section~\ref{sec:lambda_gen} where we directly measure its effect on performance).
The embedding matrices for encoding and decoding sequences are tied together,
and the attention mask $\alpha$  is computed using a bilinear term~\cite{chen-etal:2016}.
More extensive comparisons regarding the hyper-parameter choices are presented in Section~\ref{sec:hyperparameter}.

\subsection{Experimental Results}
We first report the results of the main neural-network models we consider, against
random-choice and surface-based baselines, see Table~\ref{table:agmc_better}.

For a given hyper-parameter configuration, the neural-network model starts from a random parameter initialization, and updates its parameters using the training set for a maximum of 1 million steps. We monitor the classification accuracies on the dev set and select the checkpoint with the optimal performance on the dev set. The selected checkpoint is then evaluated on the test set. The reported error bars on the test set are estimated based on the standard deviation of the posterior distribution of the balanced accuracy~\cite{brodersen:2010}.

\begin{table}[ht]
  \small
  \begin{center}
    \begin{tabular}{ c | c c }
      Method & Dev & Test\\ \hline
      Random-choice                   & 19.5 & 20.2 \\
      BLEU (brev-penalty=1)           & 41.9 & 42.2 \\
      Paragraph-Vectors               & 16.1 & 17.4  \\ \hline
      FFNN5                          & 57.8 & 58.1 $\pm$ 0.5 \\
      FFNN                          & 72.8 & 72.5 $\pm$ 0.4 \\
      \seqff                          & 83.5 & 83.2 $\pm$ 0.4 \\
    \end{tabular}
  \end{center}
  \caption{Classification accuracy on the \mcafp task for baselines and NN models.}
  \label{table:agmc_better}
\end{table}

The results clearly indicate the superiority of the neural-network models over the baselines.
In addition, the 5-way direct classification done by FFNN5 is inferior to
a 2-way classification approach followed by an argmax operation, a result somewhat surprising
considering that the FFNN5 model has the benefit of more information available at both train- and run-time.

The \seqff model obtains the best results, with accuracies of 83.5\% (dev) and 83.2\% (test);
this performance indicates that this architecture is superior to any of the non-recurrent ones,
and establishes a high-bar for a computer model performance on the \mcafp task.
According to the results from Table~\ref{tab:range}, this level of performance is still below the
~90\% accuracy achievable by humans, which makes \mcafp an interesting challenge for
future Machine Comprehension models.

In Table~\ref{tab:extra-examples}, we provide examples of instances that span the spectrum from easy to hard,
according to the accuracy obtained by increasingly more competitive learning approaches.
Next, we show the impact on performance of some of the decisions that we made regarding the
models and the training procedure.

\subsubsection{Order of training examples}
We compare the results between training our NN models on the organized training set of \mcafp
(e.g., with data fed to the model in the 'normal' order $\langle\tb_i^1, \ab_i\rangle, \langle\tb_i^2, \ab_i\rangle, \ldots, \langle\tb_h^1, \ab_h\rangle, \langle\tb_h^2, \ab_h\rangle, \ldots$),
compared to a dataset with the tuples randomly shuffled
(e.g., with data of the form $\langle\tb_i^1, \ab_i\rangle, \ldots, \langle\tb_h^2, \ab_h\rangle, \ldots, \langle\tb_i^2, \ab_i\rangle, \ldots, \langle\tb_h^1, \ab_h\rangle, \ldots$).
Results for this comparison are shown in Table \ref{table:shuffle}.

\begin{table}[ht]
  \small
  \begin{center}
    \begin{tabular}{c c |c c}
      Method & Shuffled & Dev & Test\\ \hline
      FFNN & no & 70.5 & 70.2 $\pm$ 0.4 \\
      FFNN & yes & 72.8 & 72.5 $\pm$ 0.4\\ \hline
      \seqff & no & 80.4 & 79.4 $\pm$ 0.4 \\
      \seqff & yes & 83.5 & 83.2 $\pm$ 0.4 \\
    \end{tabular}
  \end{center}
  \caption{The impact of dataset-shuffling on \mcafp accuracy.}
  \label{table:shuffle}
\end{table}

We see that for both models, the randomly-shuffled dataset performs better than the organized dataset.
The results seem to indicate that our models have their parameters tuned better when the training signal
is not presented in contiguous chunks (five binary decisions about a given $\ab_i$), and instead
is presented as independent binary decisions for various $\ab_i$ and $\ab_h$s.

\subsubsection{Difficulty of training examples}
We also wanted to understand the extent to which the difficulty of the
decoy answers influences the model's ability to solve the \mcafp task.
More precisely, we want to answer these two questions:
is using 'difficult' decoy answers always the best strategy when training such an MC model?
would a model trained on both 'easy' and 'difficult' decoys
converge to a better optimum?

To that end, we also created an additional dataset where the decoy targets are easier:
the four decoy targets for a specific source $\ab_i$ are chosen randomly from the candidate targets for all sources.
Let us denote this version of the dataset as \mcafpr.
We keep constant the dev and test splits for \mcafp, and allow the training dataset to be produced as \mcafp, \mcafpr,
as well as a combination of both (with one true and eight decoy answers).
The results of using these setups are shown in Table \ref{table:random_decoy}.

\begin{table}[ht]
  \small
  \begin{center}
    \begin{tabular}{c c |c c}
      Method & Train & Dev & Test\\ \hline
      FFNN & {\scrs \mcafp}             & 72.8 & 72.5 $\pm$ 0.4\\
      FFNN & {\scrs \mcafpr}            & 62.0 & 61.0 $\pm$ 0.5\\
      FFNN & {\scrs \mcafp + \mcafpr}   & 75.5 & 75.8 $\pm$ 0.4\\ \hline
      \seqff & {\scrs \mcafp}           & 83.5 & 83.2 $\pm$ 0.4 \\
      \seqff & {\scrs \mcafpr}          & 73.1 & 73.3 $\pm$ 0.4 \\
      \seqff & {\scrs \mcafp + \mcafpr} & 81.8 & 81.4 $\pm$ 0.4 \\
    \end{tabular}
  \end{center}
  \caption{The impact of training-example--difficulty on \mcafp accuracy.}
  \label{table:random_decoy}
\end{table}
The results indicate that indeed, for certain NN architectures such as FFNN, the models can get significantly better
by learning from both 'easy' and 'difficult' examples (75.8 on test, versus 72.5 on the same test set but using only 'difficult' examples in training).
Since we keep fixed the dev and test datasets to the ones from \mcafp, it is interesting to
see the FFNN model benefit from being trained on a dataset with a different distribution than the one on which is being tested.

This effect, however, is not observed for the \seqff model.
This model does not benefit from training on 'easy' examples, and the model resulting from training on \mcafp + \mcafpr
has an accuracy level of 81.4 (test) that is statistically-significant lower than the best result of 83.2 (test).

\subsection{Hyper-parameters for \seqff}
\label{sec:hyperparameter}
In this section, we investigate the effect of various hyper-parameter settings on the performance of the \seqff model.

\subsubsection{Effect of $\lambda_{gen}$}
\label{sec:lambda_gen}
We compare models with different values of $\lambda_{gen}$, affecting the loss-function $L(\thetab, \wb, \ub)$
defined in Equation~\ref{eq:mloss}.
As we can see from Table~\ref{table:lambda_gen}, lower $\lambda_{gen}$ values lead to higher MC accuracy scores.
This observation agrees with the intuition that training the model using a loss $L(\thetab, \wb, \ub)$ with larger $\lambda_{gen}$ means the MC loss (the first term of the loss)
may get overwhelmed by the word-generation loss (the second term).
On the other hand, higher $\lambda_{gen}$ values trigger an increase in performance associated with the generation task,
as measured by the ROUGE-L~\cite{lin-och:2004} score between the generated answer and the ground-truth answer (last column in Table~\ref{table:lambda_gen}).

\begin{table}[htp]
  \small
  \begin{center}
    \begin{tabular}{c | c c| c}
      $\lambda_{gen}$ & Dev & Test & ROUGE-L\\ \hline
      0.0 &  83.2 & 82.7 $\pm$ 0.4 & 0.1\\
      0.01&  83.5 & 83.2 $\pm$ 0.4 & 0.7\\
      0.1 &  82.8 & 83.0 $\pm$ 0.4 & 11.4 \\
      0.5 &  82.8 & 82.3 $\pm$ 0.4 & 27.7\\
      1.0 &  76.9 & 76.4 $\pm$ 0.4 & 27.9\\
      2.0 &  78.1 & 77.3 $\pm$ 0.4 & 33.2\\
      5.0 &  79.3 & 78.7 $\pm$ 0.4 & 34.5\\
      10.0 & 75.0 & 74.2 $\pm$ 0.4 & 35.1\\
    \end{tabular}
  \end{center}
  \caption{The impact of $\lambda_{gen}$ on \mcafp accuracy (as well as generation accuracy).}
  \label{table:lambda_gen}
\end{table}
The conclusion we derive from the results in Table~\ref{table:lambda_gen} is that the \seqff model is capable of tuning both
its encoding and decoding mechanisms according to the loss, which directly translates into the corresponding level of MC accuracy.
For $\lambda_{gen}=0.01$, the loss favors directly the classification task, resulting on high accuracy on both dev (83.5) and test (83.2).
On the other hand, a setting like $\lambda_{gen}=5.0$ gives a reasonable MC accuracy (78.7 on test, significantly higher than the best
obtained by any FFNN model), while also producing high-scoring abstractive answers (34.5 ROUGE-L score).
This is akin to a situation in which a student makes a multiple-choice decision by also producing, in addition to an answer choice 1-5, their own original answer.
For instance, for the last example in Table~\ref{tab:mcafp-examples}, this model generates from the 'Article' entry the following abstractive answer:
'US envoy says Israel-Syria peace talks to resume.'

\subsubsection{Effect of tied-embeddings and bilinear-attention}
\label{sec:ablation-tied}
We also compare the impact on the MC accuracy for the \seqff model of
several choices regarding the embedding matrices, as well as the implementation of the attention model.
The results are shown in Table~\ref{table:tied_bilinear} (using a fixed $\lambda_{gen}=0.01$).

\begin{table}[htp]
  \small
  \begin{center}
    \begin{tabular}{c c | c c}
      Tied-Embeddings & Bilinear-Attn & Dev & Test\\ \hline
      no & no  & 70.7 & 70.2 $\pm$ 0.4 \\
      yes & no  & 75.7 & 74.6 $\pm$ 0.4 \\
      no & yes  & 77.9 & 77.1 $\pm$ 0.4 \\
      yes & yes & 83.5 & 83.2 $\pm$ 0.4 \\
    \end{tabular}
  \end{center}
  \caption{The impact of tied-embeddings and bilinear attention on the \seqff model.}
  \label{table:tied_bilinear}
\end{table}

The results indicate that tied embeddings have a net positive effect:
it reduces the number of model parameters by nearly half,
while also contributing between 4 and 6 absolute points in accuracy.
The mechanism of bilinear attention~\cite{chen-etal:2016} is also superior to the one using only a tanh layer~\cite{bahdanau-etal:2015},
by influencing the decoder outputs such that the loss favoring classification accuracy ($\lambda_{gen}=0.01$) is minimized.
It contributes about 5 absolute accuracy-points, in a manner that is orthogonal with the contribution of the tied-embeddings,
for a high-water--mark of 83.2\% accuracy on the \mcafp test set.

\section{Conclusion}
The task of making machine truly comprehend written text is far from solved.
To make meaningful progress, we need to attack the problem on two different fronts:
build high-quality, high-volume datasets for training and testing machine-comprehension models;
and, given the full complexity of natural language~\cite{winograd:1972},
invent comprehension models that can be trained to perform well on this task.

This paper presents contributions on each of these fronts.
The algorithm presented in Section~\ref{sec:alg}, \textsc{MC-Create},
can be used to create datasets that are both high quality and high volume,
starting from data that appears naturally in today's world (e.g., news).
Since the method presented is language-agnostic, it can also be used to create
MC datasets in languages other than English, a promising prospect.
Regarding models of comprehension, the proposed hybrid \seqff model not only
performs at a high-level on the \mcafp dataset, but has the potential to
tackle a lot of other tasks that present both a generation and a discrimination component.
Such advances in the machine-comprehension task have the potential to move the field forward
in the pursuit of genuine understanding of natural language by computers.


\newpage
\begin{table*}[t]
  \begin{center}
    \begin{tabular}{l | l } \hline\hline
Correct & Incorrect \\ \hline
\multicolumn{2}{l}{BLEU (brev-penalty=1)} \\ \hline
\sm The vice-mayor of Beijing in charge of traffic management & \sm Somali pirates on Saturday captured a Thai bulk carrier with \\
\sm has resigned and been sent to the far-western region of & \sm its 27 crew members in the Arabian Sea, a maritime official \\
\sm Xinjiang as officials take drastic steps to ease chronic gridlock & \sm said. The Thor Nexus was seized in the early hours while on \\
\sm in the Chinese capital. Huang Wei's resignation and      & \sm its way to Bangladesh from the United Arab Emirates. All its \\
\sm appointment as vice-chairman of Xinjiang were approved   & \sm crew members are Thai, said Andrew Mwangura of the East \\
\sm Thursday, the official Xinhua news agency said -- the same day & \sm African Seafarers Assistance Programme. \\
\sm Beijing announced plans to slash the number of new cars in & \sm \\
\sm the city next year. \\
\\
\sm 1. Argentine official resigns amid economic crisis & \sm 1. Somali pirates seize Panama freighter: official \\
\sm {\bf\em 2. Beijing traffic official resigns amid gridlock woes} & \sm 2. {\em Somali pirates seize Spanish fishing boat: maritime official}\\
\sm 3. Italian PM Berlusconi's party in crisis as top official resigns & \sm 3. Somali pirates seize Thai fishing vessels, 77 crew: EU force \\
\sm 4. Sri Lanka budget raises fears of inflation, top official & \sm 4. {\bf Somali pirates seize Thai vessel, 27 crewmen: official} \\
\sm \hspace{3mm} resigns                                        & \sm 5. Somali pirates seize two more cargo vessels: official \\
\sm 5. Top Dhaka University official resigns amid fresh campus \\
\sm \hspace{3mm} violence & \\ \hline
\multicolumn{2}{l}{\seqff} \\ \hline
\sm Yemen has deployed new anti-terrorism forces in the country's & \sm As almost round-the-clock night descends on their country in \\
\sm restive south, the ministry of interior announced Saturday, as & \sm  December, Swedes turn to an informal celebration of light, \\
\sm Washington urged Sanaa to step up its fight against          & \sm  in keeping with long-running traditions and an effort to cope \\
\sm Al-Qaeda. The announcement follows a spate of deadly attacks & \sm with the darkest time of the year. Candles appear in the \\
\sm on government and military targets in the south, the latest & \sm windows of homes, shops, offices and cafes throughout Sweden\\
\sm on Friday when according to a security official a suspected & \sm from the start of Advent, on the fourth Sunday before \\
\sm Al-Qaeda militant shot dead a soldier who ferried him on his & \sm Christmas, to the end of December, when dusk can start \\
\sm motorbike taxi in Zinjibar , capital of Abyan province.      & \sm creeping over Stockholm as early as 2:00pm.\\
\\
\sm 1. Five nations to form anti-terrorism network            & \sm 1. {\bf At the darkest time of year, Swedes make their own light} \\
\sm 2. Iraq, Sudan, Libya join Arab anti-terrorism conference & \sm 2. Gebrselaisse runs best 1,500m time of year \\
\sm 3. Pakistani police to install 100 anti-terrorism surveillance & \sm 3. {\em Making light work: The 50-year odyssey of the laser} \\
\sm \hspace{3mm} cameras                                           & \sm 4. RugbyU : Highlanders record their second win of the year\\
\sm 4. Yemen says US did not ask to participate in al-Qaeda hunt   & \sm 5. Six-month-old girl, three-year-old boy tie the knot in Nepal \\
\sm 5. {\bf\em Yemen says anti-terrorism forces sent to south} &  \\
\\ \hline
\multicolumn{2}{l}{Human Annotator} \\ \hline
\sm Organisers of South Korea's first Formula One event say they & \sm Pope Benedict XVI has stepped up his displays of contrition \\
\sm are confident they can complete the brand-new circuit before & \sm towards victims of paedophile priests on his state visit to \\
\sm a final inspection from the sport's world governing body. & \sm Britain, but he still has far to go to win back public opinion, \\
\sm India's Karun Chandhok on Saturday became the first F1 driver & \sm observers said. Benedict met five British abuse victims and \\
\sm to test out the 5.6-kilometre (3.5-mile)track, which boasts Asia's & \sm expressed "deep sorrow" during mass on Saturday, in the \\
\sm longest straight stretch, and declared it basically in good shape. & \sm latest of several attempts to tackle an issue that is \\
\sm [...]                                                              & \sm rocking the Catholic Church [...].\\
\\
\sm 1. Formula One: SKorea to build Asia's longest track & \sm 1. Brazil's indigenous people still victims of abuse: report\\
\sm 2. Formula One: India to hold F1 race in 2009 & \sm 2. Pope's Australia sex abuse apology not enough: critics \\
\sm {\bf\em 3. Formula One: Builders race to finish first S.Korea F1 track} & \sm {\bf 3. Pope's abuse scandal apologies still not enough: experts} \\
\sm 4. Formula One: India to hold F1 race in 2010 & \sm {\em 4. Pope's sex abuse apology not enough, some victims say} \\
\sm 5. Formula One: S.Korea's F1 track to be ready in late August  & \sm 5. Vatican's toughening of abuse rules is only a start: experts \\
\hline\hline
    \end{tabular}
  \end{center}
  \caption{Examples from the dev set of the \mcafp dataset. The options in {\bf bold} are the groundtruth answers,
whereas the options in {\em italics} are the selected answers (by the respective methods). The left-side column shows
cases where the respective methods got the correct answers, while the right-side column shows cases where the answers
were incorrect.}
  \label{tab:extra-examples}
\end{table*}

\end{document}